\documentclass[11pt]{article}
\usepackage[preprint]{acl}

\usepackage{times}
\usepackage{latexsym}

\usepackage{url}            
\usepackage{booktabs}       
\usepackage{amsfonts}       
\usepackage{nicefrac}       
\usepackage{microtype}      
\usepackage[table]{xcolor}   
\usepackage{latexsym}
\usepackage{amssymb}
\usepackage{amsmath}
\usepackage{booktabs}
\usepackage{enumerate}
\usepackage{graphicx}
\usepackage{subfigure}
\usepackage{xspace}
\usepackage{float}
\usepackage{bbm}
\usepackage{bm}
\usepackage{multirow}
\usepackage{booktabs}
\usepackage{color}
\usepackage{framed}
\usepackage{stfloats}
\usepackage{iitem}
\usepackage{amssymb}
\usepackage{pifont}
\usepackage{arydshln}
\usepackage{enumitem}
\usepackage{wrapfig}
\usepackage{algorithm}
\usepackage{algpseudocode}
\usepackage{array}
\usepackage{tabularx}
\usepackage{booktabs}
\usepackage{float}
\usepackage{cleveref}
\usepackage{colortbl}
\usepackage{etoolbox}
\usepackage{tcolorbox}

\newcommand{\baserow}{\rowcolor{blue!10}}
\newcommand{\basecell}{\cellcolor{blue!10}}

\newcommand{\openrow}{\rowcolor{white!10}}

\newcommand{\ourrow}{\rowcolor{white!10}}
\newcommand{\ourcell}{\cellcolor{white!10}}

\usepackage[T1]{fontenc}

\usepackage[utf8]{inputenc}

\usepackage{microtype}

\usepackage{inconsolata}

\usepackage{graphicx}

\newcommand{\method}{{MathAgent}\xspace}
\title{\method: Adversarial Evolution of Constraint Graphs for \\Mathematical Reasoning Data Synthesis}

\author{
 \textbf{Zixiong Yu\textsuperscript{1,2},~}
 \textbf{Jun Rao\textsuperscript{3}\thanks{Co-first Author.},~} 
 \textbf{Guhan Chen\textsuperscript{2},~} 
 \textbf{Songtao Tian\textsuperscript{2},~}\\
 \textbf{Bohan Li\textsuperscript{2,4},~} 
 \textbf{Jiansheng Wei\textsuperscript{1},~} 
 \textbf{Min Zhang\textsuperscript{3},~} 
 \textbf{and Xiaojun Meng\textsuperscript{1}\thanks{Corresponding Author.}}
\\
 \textsuperscript{1}Huawei Large Model Data Technology Lab\quad
 \textsuperscript{2}Tsinghua University\\
 \textsuperscript{3}Harbin Institute of Technology, Shenzhen\quad
 \textsuperscript{4}Kyoto University
\\
\texttt{\{yuzx19,tiansongtao.2020,libh19\}@tsinghua.org.cn}\\
\texttt{rao7jun@gmail.com}\quad\texttt{chen-gh23@mails.tsinghua.edu.cn}\\
\texttt{zhangmin2021@hit.edu.cn}\quad\texttt{\{weijiansheng,xiaojun.meng\}@huawei.com}
}

\begin{document}
\maketitle
\begin{abstract}
Synthesizing high-quality mathematical reasoning data without human priors remains a significant challenge. Current approaches typically rely on seed data mutation or simple prompt engineering, often suffering from mode collapse and limited logical complexity. This paper proposes a hierarchical synthesis framework that formulates data synthesis as an unsupervised optimization problem over a constraint graph followed by semantic instantiation, rather than treating it as a direct text generation task. We introduce a \textit{Legislator-Executor} paradigm: The \textit{Legislator} adversarially evolves structured generation blueprints encoding the constraints of the problem, while the \textit{Executor} instantiates these specifications into diverse natural language scenarios. This decoupling of skeleton design from linguistic realization enables a prioritized focus on constructing complex and diverse logical structures, thereby guiding high-quality data synthesis. Experiments conducted on a total of 10 models across the Qwen, Llama, Mistral, and Gemma series demonstrate that our method achieves notable results: models fine-tuned on 1K synthesized samples outperform widely-used datasets of comparable scale (LIMO, s1K) across eight mathematical benchmarks, exhibiting superior out-of-distribution generalization.
\end{abstract}
\section{Introduction}
In recent years, Large Language Models (LLMs; \citealp{vaswani2017attention,brown2020language,zhao2026surveylargelanguagemodels}) have become a central pillar of modern artificial intelligence (AI). Although theoretical understanding of their underlying mechanisms remains relatively limited \citep{jacot2018neural, li2024eigenvalue, yu2025divergence}, LLMs have demonstrated strong reasoning abilities in practice and achieved remarkable success on complex tasks \citep{wei2022emergent,wei2026mirage}. These capabilities have in turn enabled rapid expansion into a wide range of domains, such as embodied AI \citep{driess2023palm,zeng2026janusvln} and LLM-driven scientific discovery \citep{boiko2023autonomous,lu2024aiscientistfullyautomated}.
This progress has been driven by multiple factors, including the scaling of model parameters and training data \citep{kaplan2020scalinglawsneurallanguage, hoffmann2022trainingcomputeoptimallargelanguage}, reasoning-oriented techniques such as chain-of-thought (CoT) prompting \citep{wei2022chain,zeng2025futuresightdrive}, and, equally importantly, the quality of training data \citep{zhou2023lima,ye2025limo,zhao2026kalmembeddingv}. 
However, as high-quality human-generated corpora become increasingly difficult to scale, the field faces a growing data bottleneck \citep{villalobos2024position}. Consequently, synthetic data generation, which uses generative models to produce training samples, has emerged as a major research direction \citep{honovich-etal-2023-unnatural,rao-etal-2025-SEAPO,ke-etal-2025-aquilt}.

Current synthesis paradigms primarily fall into two categories: (i) \textit{Seed-based} methods, such as Self-Instruct \citep{wang-etal-2023-self-instruct}, expand upon human-curated seeds. While effective, their diversity is inherently upper-bounded by the semantic span of the initial seeds. (ii) \textit{Zero-shot} methods like Magpie \citep{xu2025magpie} probe model distributions directly but often suffer from mode collapse and logical hallucinations due to the lack of structural guidance \citep{shumailov2023curse}. We argue that by framing data synthesis as a mere text generation task rather than a structured optimization problem, current methods often confine models to superficial narrative imitation without mastering core reasoning capabilities \citep{gudibande2023false}.


To address this, we propose a hierarchical synthesis framework anchored by a bi-level \textit{Legislator-Executor} paradigm, which effectively decouples structural specifications from their textual instantiation. By pre-establishing high-level task blueprints that incorporate logical relations and constraints, the framework more effectively guides the generation of high-quality mathematical problems. We instantiate this architecture as \method, where the Legislator (meta-level) adversarially optimizes the combination of problem elements over a constraint graph, while the Executor (base-level) transforms these abstract blueprints into natural language.

This decoupling mechanism enables a prioritized focus on orchestrating structural diversity and complexity. Through iterative adversarial evolution, {\method} continuously explores the underlying structural space, thereby progressively pushing the frontiers of model generation capabilities. Compared to direct probing methods confined by high-frequency patterns and seed dataset augmentation methods limited by initial semantic ranges, our framework excels at capturing scarce data characterized by high difficulty and quality. Consequently, relying solely on basic conceptual primitives rather than seed data, \method synthesizes corpora with high structural complexity and rich diversity, while flexibly regulating the complexity of data distributions through an adaptive early-stop iteration mechanism.


Our contributions are summarized as follows:
\begin{itemize}[leftmargin=*,nosep]
    \item We propose the Legislator-Executor paradigm, a hierarchical synthesis framework that decouples task specification from textual realization to facilitate the guided synthesis of reasoning data.

    \item We introduce a constraint-graph-based adversarial evolutionary mechanism to explore structural spaces, generating high-difficulty, high-quality problems often absent in standard datasets.

    \item Extensive experiments demonstrate that models fine-tuned on 1K \method samples outperform mainstream datasets of comparable scale (LIMO \& s1K) across eight benchmarks, exhibiting superior out-of-distribution generalization.
\end{itemize}

\section{Related Work}

\paragraph{Data Synthesis} 
A prevalent paradigm in data synthesis involves the iterative expansion of seed examples. 
Methods like Self-Instruct \citep{wang-etal-2023-self-instruct} and WizardMath \citep{luo2025wizardmath} utilize evolution strategies to amplify task complexity, while MathGenie \citep{lu-etal-2024-mathgenie} employs a backward mechanism, augmenting seed solutions to back-translate new questions. Nevertheless, these approaches are constrained by the "semantic radius" of their initial seeds, often failing to explore the unknown regions of the problem space. 
Similarly, for preference-pair data, methods such as SeaPO \citep{rao-etal-2025-SEAPO} construct preference pairs by generating contrastive responses based on existing answers; however, such approaches typically place higher demands on the fine-grained and controllable editing capabilities of LLMs \citep{zeng-etal-2025-bridging}.
To avoid relying on seed datasets, zero-shot methods such as Magpie \citep{xu2025magpie} and schema-driven frameworks like Condor \citep{maosongcao-etal-2025-condor} attempt to synthesize data from scratch. While successful in high-resource domains, they lack the structural incentives to discover samples in the long-tail distribution, where complex reasoning capabilities are often forged. 

\paragraph{Multi-Agent and Adversarial Generation}
The deployment of LLMs has progressed from single-turn prompting to complex Multi-Agent Systems. Frameworks such as CAMEL \citep{li2023camel} and MetaGPT \citep{hong2024metagpt} demonstrate that role-playing agents can effectively decompose tasks through cooperation, while AgentDropout \citep{wang-etal-2025-agentdropout} further improves efficiency and coordination via dynamic agent elimination.
In the realm of data synthesis, MATRIX \citep{tang2024synthesizing} utilizes multi-agent simulation to construct virtual societies, generating instruction data grounded in realistic social scenarios. 
Concurrently, multi-agent debate \citep{du2024improving} has emerged as a pivotal mechanism for enhancing reasoning reliability. For instance, Debate4MATH \citep{luo2024debate4math} employs fine-grained step verification to rectify logical errors, while \citet{liang-etal-2024-encouraging} leverage debate to stimulate divergent thinking for higher-quality problem solving. 
Our work adapts adversarial dynamics for data synthesis, 
aiming to drive continuous evolution of the training data distribution and explore the generation of complex samples.

\section{Method}
\label{sec:method}
We propose a hierarchical synthesis framework to synthesize mathematical data by optimizing the constraint graph of problem structures. 
Specifically instantiated as {\method}, our approach diverges from standard methods that operate directly in the token space by decoupling the synthesis process into two distinct phases: (1) Structural Evolution (Meta-level), governed by a \textit{Legislator} agent that optimizes a constraint graph (acting as the synthesis blueprint); and (2) Semantic Instantiation (Base-level), conducted by an \textit{Executor} that grounds the graph structure into natural language scenarios.

\subsection{Problem Formulation}
We formulate the skeleton of a mathematical problem as a structure comprising a \textbf{Constraint Graph} $\mathcal{G} = (\mathcal{V}, \mathcal{E})$ and \textbf{Style Tokens} $\mathcal{S}$. Specifically:
\begin{itemize}[leftmargin=*,nosep]
    \item \textbf{Nodes ($\mathcal{V}$)} represent mathematical concepts.
    \item \textbf{Edges ($\mathcal{E}$)} represent logical relations.
    \item \textbf{Style Tokens ($\mathcal{S}$)} control global attributes (e.g., problem category or difficulty level).
\end{itemize}
Serving as a blueprint for problem synthesis, this graph facilitates automated structural evolution, enabling the targeted generation of high-complexity and richly diverse reasoning data.

Formally, our objective is to explore the space of graph topologies, optimizing the complexity while ensuring strict solvability (conditioned on $\mathcal{S}$):
\begin{align*}
\mathcal{G}^* = \text{arg\,max}_{\mathcal{G} \in \mathbb{G}}\;\mathcal{H}(\mathcal{G})\quad\text{s.t.}\quad\mathbb{I}_{\text{valid}}(\mathcal{G}\mid\mathcal{S}) = 1,
\end{align*}
where $\mathbb{G}$ is the search space, $\mathcal{H}(\cdot)$ estimates the complexity, and $\mathbb{I}_{\text{valid}}(\cdot)$ is a binary validity indicator. Note that while the optimization seeks to push the reasoning frontier (finding $\mathcal{G}^*$), the evolutionary trajectory yields a diverse curriculum of graphs.
The resulting tuple $(\mathcal{G}^*, \mathcal{S})$ is then passed to the Executor for textual realization. 

\begin{figure*}[t]
    \centering
    \includegraphics[width=0.99\linewidth]{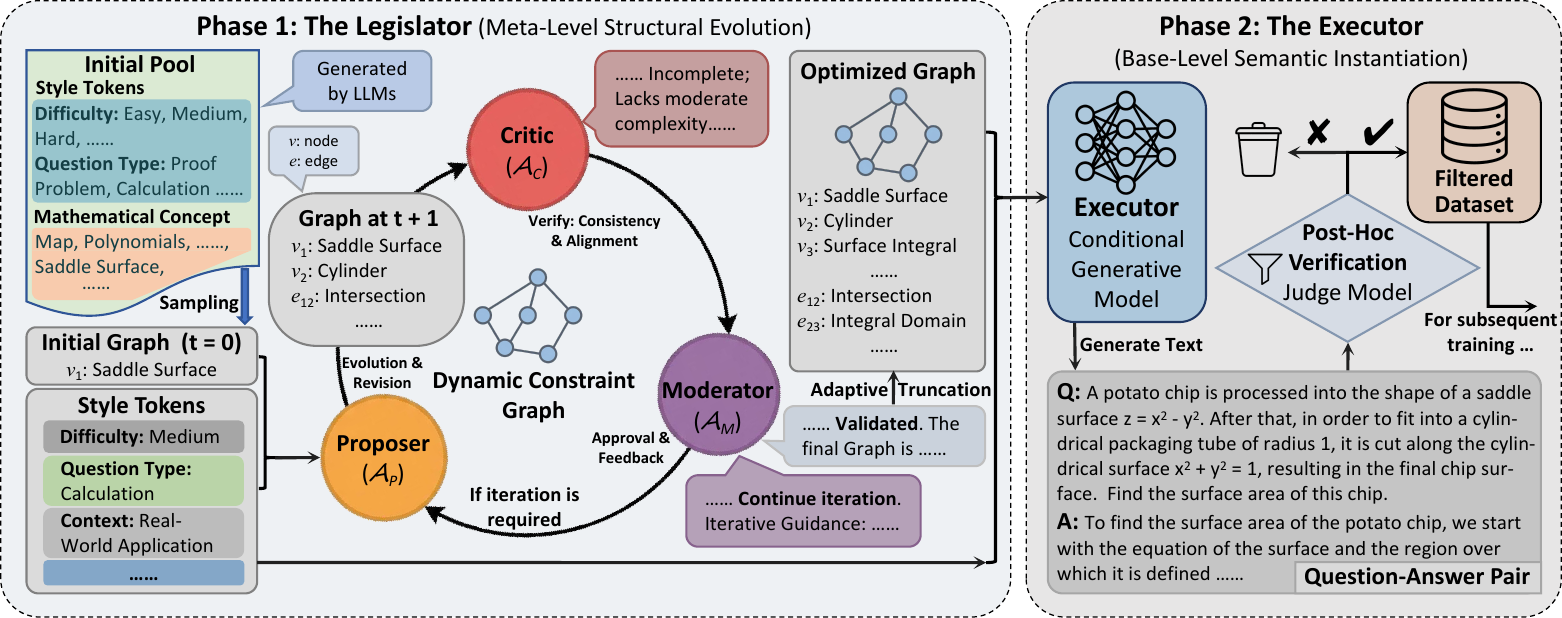}
    \caption{\textbf{The \method Framework.} The framework consists of two decoupled phases: (1) Meta-Level Structural Evolution, where a tri-agent Legislator system (Proposer, Critic, and Moderator) iteratively optimizes a Constraint Graph $\mathcal{G}$ based on Style Tokens $\mathcal{S}$; and (2) Base-Level Semantic Instantiation, where the Executor grounds the optimized structural blueprint into natural language problems $Q$ and reasoning chains $A$.}
    \label{fig: main_figure}
\end{figure*}
\subsection{Phase 1: The Legislator (Meta-Level)}
To address the optimization objective, we design the Legislator as a tri-agent evolutionary system. Instead of directly manipulating text, the system iteratively optimizes the dynamic constraint graph $\mathcal{G}_t$ through inter-agent collaboration under given style token $\mathcal{S}$ conditions. The evolutionary process is jointly driven by three distinct roles:

\paragraph{Proposer ($\mathcal{A}_P$)} As the driving engine of structural evolution, the proposer $\mathcal{A}_P$ optimizes $\mathcal{G}_t$ to $\mathcal{G}_{t+1}$ guided by feedback from previous iterations. It resolves logical contradictions while ensuring the graph’s characteristics align with the structural specifications defined in style tokens $\mathcal{S}$. In particular, if the current structure has not reached the target complexity required by style tokens $\mathcal{S}$, the proposer $\mathcal{A}_P$ proactively expands knowledge nodes or strengthens constraints to enhance structural depth. 

\paragraph{Critic ($\mathcal{A}_C$)} 
As a key component of adversarial evolution, the critic $\mathcal{A}_C$ scrutinizes $\mathcal{G}_{t+1}$ across three dimensions based on the style tokens $\mathcal{S}$: (1) \textit{Internal Consistency}: it verifies whether logical contradictions exist in the current graph; (2) \textit{Specification Alignment}: it checks if the graph complies with the constraints specified in the style tokens $\mathcal{S}$; 
(3) \textit{Optimization Potential}: it proactively probes for superior configurations that transcend the current design. The results are synthesized into a comprehensive refinement report.


\paragraph{Moderator ($\mathcal{A}_M$)} 
The moderator $\mathcal{A}_M$ serves as the strategic decision-maker, adjudicating the evolution of $\mathcal{G}_{t+1}$ by weighing the refinement report against the global objective. Each cycle yields one of two outcomes:
\begin{itemize}[leftmargin=*,nosep]
    \item \textit{Adaptive Truncation}: If $\mathcal{G}_{t+1}$ satisfies $\mathcal{S}$ and the potential for further gain is marginal, $\mathcal{A}_M$ terminates the process and outputs $\mathcal{G}^*$.
    \item \textit{Iterative Guidance}: Otherwise, $\mathcal{A}_M$ directs $\mathcal{A}_P$ to implement the critic's suggestions to resolve inconsistencies or enhance structural depth.
\end{itemize}

\paragraph{Initialization}
To ensure high initial diversity and eliminate human intervention, we deploy a similar adversarial mechanism before the evolutionary loop to construct an initial pool: the proposer $\mathcal{A}_P$ activates latent information to propose candidate attributes, while the critic $\mathcal{A}_C$ filters these attributes based on requirements such as orthogonality, validity, and diversity. This adversarial process builds a self-organized initial pool, containing:

\begin{itemize}[leftmargin=*,nosep]

\item \textbf{Style Tokens ($\mathcal{S}$)}: A rich and diverse set of stylistic constraint dimensions, each offering as comprehensive a range of options as possible.

\item \textbf{Concept Taxonomy ($\mathcal{C}$)}: A comprehensive atlas of mathematical domains.

\end{itemize}
At the onset ($t=0$), the system randomly samples from this initial pool to generate the initial graph $\mathcal{G}_0$, ensuring that the entire dataset is driven solely by the model's intrinsic representational diversity. 

In practice, the concept taxonomy can be derived from a variety of sources, such as existing knowledge bases, interactions between humans and LLMs, or weaknesses identified from evaluations of the target LLM \citep{rao-etal-2025-apt}.

\subsection{Phase 2: The Executor (Base-Level)}
The Executor is a conditional generative model that performs semantic instantiation. It receives the linearized textual representation of the constraint graph $\mathcal{G}^*$ alongside the set of Style Tokens $\mathcal{S}$: 
\begin{align*}
    (Q, A) \sim P_{\text{executor}}(\,\cdot\, \mid \mathcal{G}^*, \mathcal{S})
\end{align*}
where $Q$ denotes the natural language problem statement and $A$ is the step-by-step reasoning chain.
By conditioning the generation on $\mathcal{G}^*$, the executor is freed from the burden of exploring and constructing complexity and diversity, allowing it to focus solely on language itself, thereby enabling the generation of diverse textual scenarios.

To further ensure the reliability of the synthesized question-answer pairs, we adopt a general model-based verification scheme \citep{zheng2023judging}, which involves employing an external model as a judge to evaluate the logical correctness of the generated questions and answers, as well as the consistency between their descriptions. Only samples that pass this verification are retained.

\begin{table*}[!t]
\centering
\resizebox{0.99\textwidth}{!}{%
\begin{tabular}{ll*{8}{c}c}
\toprule
\multirow{2}{*}{\textbf{\centering Model}} & \multirow{2}{*}{\textbf{Dataset}} & \multicolumn{2}{c}{\textbf{Elementary}} & \multicolumn{3}{c}{\textbf{Middle}} & \multicolumn{3}{c}{\textbf{Competition}} & \multirow{2}{*}{\parbox{1.3cm}{\centering\textbf{Avg.}}} \\
\cmidrule(lr){3-4} \cmidrule(lr){5-7} \cmidrule(lr){8-10} & 
& \parbox{1.3cm}{\centering\small\textbf{GSM8K}} 
& \parbox{1.3cm}{\centering\small{\textbf{MATH}}\\\centering\small{\textbf{500}}}
& \parbox{1.3cm}{\centering\small{\textbf{Minerva}}\\\centering\small{\textbf{Math}}} 
& \parbox{1.3cm}{\centering\small\textbf{Gaokao}\\\centering\small{\textbf{2023en}}} 
& \parbox{1.3cm}{\centering\small\textbf{Olympiad}\\\centering\small\textbf{Bench}} 
& \parbox{1.3cm}{\centering\small\textbf{AIME24}\\\centering\small\textbf{(Avg@8)}} 
& \parbox{1.3cm}{\centering\small\textbf{AIME25}\\\centering\small\textbf{(Avg@8)}} 
& \parbox{1.3cm}{\centering\small\textbf{AMC23}\\\centering\small\textbf{(Avg@8)}} & \\
\midrule
\multicolumn{11}{c}{\emph{Qwen3 Series Models}}\\
\baserow\basecell Qwen3-14B-Base & Base & 94.7 & 80.6 & 37.5 & 67.5 & 46.2 & 15.0 & 15.8 & 66.6 & 53.0 \\
\openrow & LIMO & 91.8 & 86.2 & 39.0 & 76.9 & 50.8 & 33.8 & 27.5 & 70.0 & 59.5 \\
\openrow & S1K & 87.5 & 86.4 & \textbf{40.8} & 76.1 & 52.6 & 37.9 & 25.0 & 75.9 & 60.3 \\
\ourrow\ourcell & Ours & \textbf{95.4} & \textbf{91.8} & 39.0 & \textbf{79.0} & \textbf{56.3} & \textbf{38.8} & \textbf{30.0} & \textbf{80.6} & \textbf{63.9} \\
\baserow\basecell Qwen3-8B-Base & Base & 92.0 & 76.8 & 32.7 & 64.9 & 41.8 & 17.5 & 13.8 & 58.4 & 49.7 \\
\openrow & LIMO & 88.3 & 80.4 & 35.7 & 68.6 & 44.9 & 19.6 & 24.2 & 60.3 & 52.8 \\
\openrow & S1K & 87.6 & 81.4 & 37.5 & 71.7 & 44.4 & 19.2 & 23.3 & 60.3 & 53.2 \\
\ourrow\ourcell & Ours & \textbf{93.3} & \textbf{87.2} & \textbf{39.7} & \textbf{75.8} & \textbf{50.5} & \textbf{27.1} & \textbf{25.4} & \textbf{74.7} & \textbf{59.2} \\
\baserow\basecell Qwen3-4B-Base & Base & 82.8 & 72.4 & 20.2 & 60.8 & 37.9 & 10.4 & 7.5 & 50.3 & 42.8 \\
\openrow & LIMO & 82.0 & 74.0 & 30.9 & 68.1 & 40.0 & 13.8 & 19.6 & 56.9 & 48.2 \\
\openrow & S1K & 84.8 & 76.8 & 33.5 & 68.6 & 40.6 & 16.7 & 20.0 & 53.1 & 49.3 \\
\ourrow\ourcell & Ours & \textbf{92.0} & \textbf{81.8} & \textbf{35.3} & \textbf{69.6} & \textbf{44.6} & \textbf{19.2} & \textbf{20.4} & \textbf{65.0} & \textbf{53.5} \\
\midrule
\multicolumn{11}{c}{\emph{Qwen2.5 Series Models}}\\
\baserow\basecell Qwen2.5-7B & Base & 87.4 & 63.0 & 24.3 & 56.1 & 29.0 & 5.4 & 3.8 & 36.9 & 38.2 \\
\openrow & LIMO & 88.6 & 71.4 & 29.4 & 61.8 & 34.7 & 10.0 & 14.6 & 45.9 & 44.6 \\
\openrow & S1K & 89.4 & 68.8 & 30.5 & 62.9 & 35.9 & 11.7 & 9.6 & 43.8 & 44.1 \\
\ourrow\ourcell & Ours & \textbf{90.1} & \textbf{74.6} & \textbf{30.9} & \textbf{68.3} & \textbf{38.5} & \textbf{14.2} & \textbf{15.0} & \textbf{55.9} & \textbf{48.4} \\
\baserow\basecell Qwen2.5-Math-7B & Base & 66.7 & 64.0 & 12.1 & 56.1 & 28.3 & 11.7 & 4.2 & 38.8 & 35.2 \\
\openrow & LIMO & 87.4 & 72.2 & 31.6 & 63.1 & 37.9 & 10.8 & 14.6 & 47.2 & 45.6 \\
\openrow & S1K & 87.9 & 73.2 & 33.5 & 64.2 & 35.6 & 11.7 & 11.2 & 49.7 & 45.9 \\
\ourrow\ourcell & Ours & \textbf{91.6} & \textbf{82.2} & \textbf{34.2} & \textbf{70.4} & \textbf{47.0} & \textbf{18.8} & \textbf{18.3} & \textbf{65.3} & \textbf{53.5} \\
\baserow\basecell Qwen2.5-7B-QwQ & Base & 90.4 & 75.4 & 32.7 & 66.8 & 39.6 & 17.9 & 19.6 & 55.0 & 49.7 \\
\openrow & LIMO & 90.7 & 78.4 & 32.7 & 70.4 & 44.1 & 16.2 & 20.8 & 57.5 & 51.4 \\
\openrow & S1K & 90.7 & 78.2 & 30.5 & 64.9 & 44.7 & 17.9 & 20.0 & 55.0 & 50.2 \\
\ourrow\ourcell & Ours & \textbf{91.7} & \textbf{83.8} & \textbf{33.8} & \textbf{73.5} & \textbf{50.7} & \textbf{27.5} & \textbf{21.7} & \textbf{70.6} & \textbf{56.7} \\
\midrule
\multicolumn{11}{c}{\emph{Llama, Mistral and Gemma Models}}\\
\baserow\basecell Llama-3.1-8B 
& Base & 40.9 & 13.4 & 4.8 & 15.1 & 3.1 & 0.0 & 0.0 & 4.7 & 10.3 \\
\openrow & LIMO & 66.0 & 24.6 & 6.2 & 24.9 & 6.4 & 0.0 & 0.4 & 8.8 & 17.2 \\
\openrow & S1K & 66.9 & 24.0 & 8.5 & 28.8 & 5.9 & 0.0 & \textbf{0.8} & 7.8 & 17.8 \\
\ourrow\ourcell & Ours & \textbf{67.8} & \textbf{27.6} & \textbf{9.6} & \textbf{30.9} & \textbf{6.8} & \textbf{0.8} & \textbf{0.8} & \textbf{10.9} & \textbf{19.4} \\
\baserow\basecell Llama-3.2-3B 
& Base & 25.8 & 7.4 & 2.6 & 9.1 & 2.5 & 0.0 & 0.0 & \textbf{3.8} & 6.4 \\
\openrow & LIMO & 22.8 & 8.6 & 3.3 & 11.9 & 2.7 & 0.0 & 0.0 & 2.5 & 6.5 \\
\openrow & S1K & 20.4 & 7.8 & 2.2 & 11.2 & 1.8 & 0.0 & 0.0 & 3.4 & 5.9 \\
\ourrow\ourcell & Ours & \textbf{37.4} & \textbf{10.8} & \textbf{4.0} & \textbf{17.1} & \textbf{3.3} & 0.0 & \textbf{0.4} & \textbf{3.8} & \textbf{9.6} \\
\baserow\basecell Mistral-7B-v0.3 
& Base & 18.0 & 7.4 & 2.6 & 9.4 & 2.2 & 0.0 & 0.0 & 2.5 & 5.3 \\
\openrow & LIMO & 33.1 & 12.6 & 4.0 & 14.5 & 3.0 & 0.0 & \textbf{0.4} & 2.2 & 8.7 \\
\openrow & S1K & 33.4 & 7.8 & 5.9 & 11.9 & 1.8 & 0.0 & 0.0 & 2.8 & 8.0 \\
\ourrow\ourcell & Ours & \textbf{43.5} & \textbf{13.2} & \textbf{6.6} & \textbf{15.6} & \textbf{3.4} & \textbf{0.4} & \textbf{0.4} & \textbf{3.1} & \textbf{10.8} \\
\baserow\basecell Gemma-2-9B
& Base & 54.8 & 23.4 & 8.5 & 24.7 & 6.5 & 0.4 & 0.0 & 7.5 & 15.7 \\
\openrow & LIMO & \textbf{76.6} & 40.6 & 14.7 & 38.4 & 15.7 & 2.1 & 0.8 & 20.6 & 26.2 \\
\openrow & S1K & 76.4 & 41.0 & \textbf{17.3} & 43.1 & 13.5 & 1.2 & 0.4 & 20.3 & 26.7 \\
\ourrow\ourcell & Ours & 75.9 & \textbf{43.8} & 16.2 & \textbf{43.4} & \textbf{16.0} & \textbf{2.9} & \textbf{1.2} & \textbf{23.8} & \textbf{27.9} \\
\bottomrule
\end{tabular}}
\caption{\textbf{Main Results.} Model performance comparison on eight mathematical benchmarks (scores in \%). Our synthesized dataset outperforms existing open-source baselines (LIMO and s1K) for SFT across multiple models.}
\label{tab: math_performance}
\end{table*}
\section{Experiment}
\subsection{Setup}
\paragraph{Data Synthesis Implementation} We implement {\method} through a multi-model pipeline. First, \texttt{DeepSeek-V3} \citep{deepseekai2025deepseekv3technicalreport} is employed to construct candidate pools for style tokens $\mathcal{S}$ and concept taxonomy $\mathcal{C}$, establishing a diverse initial state for meta-level evolution, with this model also responsible for generating the final solutions. Next, the intermediate cyclic adversarial evolution led by the Legislator, the semantic instantiation performed by the Executor, and the subsequent verification process are all driven by \texttt{Qwen2.5-32B-Instruct} \citep{qwen2025qwen25technicalreport}. All generation stages maintain a uniform temperature of 0.3. 
We synthesize a final corpus of 1K instances, aligning the data scale with the baseline datasets introduced in the following section to ensure a fair comparison of data efficiency.

\paragraph{Baseline}
We benchmark the proposed synthetic data against datasets curated through rigorous human-designed filtering pipelines in supervised fine-tuning (SFT). To facilitate extensive experimentation, we select two well-known small-scale open-source datasets, \textbf{LIMO} \citep{ye2025limo} and \textbf{s1K} \citep{muennighoff2025s1simpletesttimescaling} (containing approximately 0.8K and 1K samples, respectively), as representatives of such high-quality filtered data. Previous studies have indicated that the quality rather than the quantity of instruction-tuning datasets is critical \citep{zhou2023lima}, making SFT experiments at this scale sufficiently informative. Details of these datasets are provided in Appendix \ref{sec: baseline_details}.

\paragraph{Models} Our fine-tuning experiments are primarily conducted on the Qwen series of models, specifically including the \texttt{Qwen3-14B/8B/4B-Base} \citep{yang2025qwen3technicalreport} models, \texttt{Qwen2.5-7B} \citep{qwen2025qwen25technicalreport}, and \texttt{Qwen2.5-Math-7B} \citep{yang2024qwen25mathtechnicalreportmathematical}.
We also include \texttt{Qwen2.5-7B-QwQ}, which is fine-tuned from the Math-Base model on 15K QwQ samples \citep{QwQ2024}.
To ensure cross-architecture generalization, we extend our evaluation to
\texttt{Llama-3.1-8B} \citep{grattafiori2024llama3herdmodels}, \texttt{Llama-3.2-3B} \citep{meta2024llama}, 
\texttt{Mistral-7B-v0.3} \citep{jiang2023mistral7b}, 
and \texttt{Gemma-2-9B} \citep{gemmateam2024gemma2improvingopen}.
Comprehensive training details are provided in Appendix \ref{sec: prompt} and \ref{sec: Training Details}. 

\paragraph{Evaluation}
We evaluate the model's mathematical capability after SFT using eight mathematical test sets across the following three difficulty levels: 

\begin{itemize}[leftmargin=*,nosep]
    \item \textbf{Elementary (Elem.)} GSM8K \citep{cobbe2021trainingverifierssolvemath} \& MATH500 \citep{hendrycks2021measuringmathematicalproblemsolving};
    \item \textbf{Middle (Mid.)} Minerva Math \citep{lewkowycz2022solvingquantitativereasoningproblems}, Gaokao 2023en \citep{liao-etal-2024-mario} \&  Olympiad Bench \citep{he-etal-2024-olympiadbench};
    \item \textbf{Competition (Comp.)} AIME 2024 \& 2025 \citep{aime} and AMC23 \citep{amc}.
\end{itemize}
For Elementary and Middle-level sets, we report greedy decoding accuracy in a zero-shot setting. For Competition-level sets, we perform 8 sampling iterations per problem and report the average accuracy to mitigate variance.  During answer generation for these datasets, we set the temperature to 0.1 and top\_p to 0.95. A maximum generation length of 8192 tokens is applied across all test sets.

\subsection{Main Results}
\paragraph{Superior Performance} 
As presented in \Cref{tab: math_performance}, our synthesized dataset yields significant mathematical reasoning improvements across a comprehensive range of model architectures (Qwen, Llama, Mistral, Gemma), scales (3B--14B), and initialization stages (Base, Math, and SFT). Notably, our method consistently outperforms competitive open-source baselines (LIMO and s1K) across all eight benchmarks, highlighting the efficacy of the Legislator-Executor paradigm. Crucially, these gains are not a byproduct of data contamination; the synthesis process remains entirely independent of the evaluation benchmarks, as substantiated by the similarity analysis in Appendix \ref{sec: Training-Test Set Similarity}.

\paragraph{Cross-Difficulty Performance}
As shown in \Cref{tab: math_performance}, the performance gains of our method are most prominent on high-difficulty benchmarks. For a clearer comparison, we selected representative models (The Mistral and Llama series were excluded due to their consistently poor performance on challenging test sets, while the Qwen3 series post-training versions were omitted as they incorporate thinking modes that preclude direct comparability) and compared them with their official instruction-tuned (Instruct) versions, 
with detailed results presented in \Cref{tab: instruct}. 

While Instruct models maintain a slight edge on elementary benchmarks (potentially by capitalizing on high-frequency linguistic patterns within their large-scale training sets), our method significantly outperforms them on intermediate and competition-level tasks. This underscores the efficacy of adversarial evolution in capturing complex skeletons and logical dependencies, fostering robust reasoning capabilities that transcend simple pattern matching.

\begin{table}[t]
\centering
\setlength{\tabcolsep}{1mm}
\resizebox{0.48\textwidth}{!}
{
\begin{tabular}{llcccc}
\toprule
\textbf{Model} & \parbox{1cm}{\centering\textbf{Method}} & \parbox{1cm}{\centering\textbf{Elem.}} & \parbox{1cm}{\centering\textbf{Mid.}} & \parbox{1cm}{\centering\textbf{Comp.}} & \parbox{1cm}{\centering\textbf{Avg.}} \\
\midrule
\multirow{2}{*}{\parbox{3cm}{Qwen2.5-7B\\\small{\citep{qwen2025qwen25technicalreport}}}} 
& Instruct & 84.4 & 45.8 & 24.4 & 47.4\\
& Ours & 83.3 & 45.9 & 28.4 & 48.7\\
\midrule
\multirow{2}{*}{\parbox{3cm}{Qwen2.5-Math-7B\\\small{\citep{yang2024qwen25mathtechnicalreportmathematical}}}} 
& Instruct & 89.5 & 48.5 & 28.6 & 51.3\\
& Ours & 86.9 & 50.5 & 34.1 & 53.5\\
\midrule
\multirow{2}{*}{\parbox{3cm}{Gemma-2-9B\\\small{\citep{gemmateam2024gemma2improvingopen}}}} 
& Instruct & 61.1 & 21.8 & 6.0 & 25.7 \\
& Ours & 59.9 & 25.2 & 9.3 & 27.9 \\
\bottomrule
\end{tabular}
}
\caption{Performance Comparison of Our Method versus the Official instruction-tuned Version. Superior performance of our method on more challenging test sets.}\label{tab: instruct}
\end{table}

\subsection{Ablation Study}
To verify the necessity of the core components, we conduct an ablation study using \texttt{Qwen2.5-7B} as the base model and \texttt{Qwen2.5-Math-72B-Instruct} \citep{yang2024qwen25mathtechnicalreportmathematical} as the resource-efficient solution annotator. The experimental configurations and results are summarized in \Cref{tab: Ablation Study}.
\begin{table}[h!]
\centering
\resizebox{0.48\textwidth}{!}{
\begin{tabular}{lcc}
\toprule
\textbf{Method / Variant} & \textbf{Avg.} & \textbf{$\Delta$} \\
\midrule
\rowcolor{gray!10} \textbf{Full \method} & \textbf{45.4} & - \\
\midrule
\textit{Impact of Legislator-Executor paradigm} & & \\
\quad w/o Constraint Graph (Direct Gen.) & 42.4 & {-3.0} \\
\midrule
\textit{Impact of Adversarial Evolution} & & \\
\quad w/o Roundtable (One-pass) & 43.1 & {-2.3} \\
\bottomrule
\end{tabular}
}
\caption{\textbf{Ablation Study.} $\Delta$ indicates the performance drop relative to the full framework. The results underscore the necessity of both structural decoupling and adversarial evolution for high-quality synthesis.}\label{tab: Ablation Study}
\end{table}

The performance degradation observed in both variants highlights the synergy between our structural and evolutionary components. Eliminating the constraint graph results in a substantial drop (-3.0\%), as the model reverts to superficial narrative imitation and high-frequency patterns in the absence of an explicit structural blueprint. Similarly, bypassing adversarial evolution (-2.3\%) forces a reliance on initial semantic intuition, confirming that continuous structural refinement and difficulty-stretching via adversarial mechanisms are essential for pushing the boundaries of the model's generative capacity for high-quality problem synthesis.

\subsection{Isolating Problem Quality}
\label{sec: Problem Quality}
In this section, we disentangle the impact of problem quality from answer generation to verify that the performance gains in \Cref{tab: math_performance} are primarily driven by our synthesis approach.

\paragraph{SFT with Consistent Response Generation} 
To eliminate the impact of differing response generation methodologies across datasets, we adopt a uniform response generation protocol. Specifically, all responses are generated in a single pass using the \texttt{Qwen2.5-Math-72B-Instruct} model without post-generation filtering.
For the baseline problem datasets, in addition to \textbf{LIMO} and \textbf{s1K} datasets already used in \Cref{tab: math_performance}, we included two additional datasets: \textbf{NuminaMath} \citep{li2024numinamath} and \textbf{Magpie} \citep{xu2025magpie}. Detailed descriptions of these datasets are provided in Appendix \ref{sec: baseline_details}.

We conducted SFT on the \texttt{Qwen2.5-7B} model using response-standardized versions of all datasets to ensure a rigorous and controlled comparison. The results in \Cref{tab: Problem Quality} reveal that datasets relying on rigorous heuristic curation (LIMO and s1K) outperform both the unfiltered NuminaMath and the unconstrained, single-prompt synthesis of Magpie. In addition, even after standardizing response generation, our proposed method maintains its superiority across all comparisons. 


\begin{table}[h!]
\centering
\resizebox{0.48\textwidth}{!}
{
\begin{tabular}{lcrc}
\toprule
\textbf{Method} & \textbf{Type} & \textbf{Size} & \textbf{Avg.} \\
\midrule
Qwen2.5-7B \small{\citep{qwen2025qwen25technicalreport}} & - & - & 38.2 \\
~~+~NuminaMath \small{\citep{li2024numinamath}} & SFT & 1K & 41.1\\
~~+~Magpie \small{\citep{xu2025magpie}} & SFT & 1K & 41.5  \\
~~+~LIMO \small{\citep{ye2025limo}} & SFT & 0.8K & 43.2 \\
~~+~S1K \small{\citep{muennighoff2025s1simpletesttimescaling}} & SFT & 1K & 43.0 \\ 
~~+~\method~(Ours) & SFT & 1K & 45.4  \\
\midrule
\multicolumn{2}{l}{Qwen2.5-7B-Instruct \small{{\citep{qwen2025qwen25technicalreport}}}} & - & 47.4\\
~~+~NuminaMath \small{\citep{li2024numinamath}} & IDPO & 32K & 47.9\\
~~+~Magpie \small{\citep{xu2025magpie}} & IDPO & 32K & 48.0\\
~~+~\method~(Ours) & IDPO & 32K & 49.1\\
\bottomrule
\end{tabular}
}
\caption{\textbf{Isolating Problem Quality.} By standardizing response generation for all datasets, we observe consistent gains under both SFT and IDPO. This confirms that the superiority of \method~stems from the inherent logical quality of the synthesized problems, independent of response-level variance.} \label{tab: Problem Quality}
\end{table}

\paragraph{IDPO} We also adopt the Iterative Direct Preference Optimization\footnote{\url{https://github.com/RLHFlow/Online-DPO-R1}} (IDPO; \citealp{zhangonline,rao2025dynamicsamplingadaptsiterative}) method for evaluation. This approach requires the model to autonomously explore the solution space during training, thereby effectively distinguishing the intrinsic quality differences among various problem sets. The experiments are conducted using the \texttt{Qwen2.5-7B-Instruct} as the base model, and the LIMO and s1K datasets are excluded from this comparison due to their limited scale. Results in \Cref{tab: Problem Quality} show that despite the limited headroom for improvement in an already instruction-tuned base model, our method still attains the superior performance. 

\subsection{Cross-Task Generalization}\label{sec: Cross-Task}
Fine-tuning on domain-specific data often incurs a trade-off in general capability degradation. To verify that our method maintains robust cross-task generalization while enhancing mathematical reasoning, we extend our evaluation to the following benchmarks: BBH \citep{suzgun2023challenging}, HumanEval \citep{chen2021evaluatinglargelanguagemodels}, MMLU \citep{hendrycks2021measuring}, and TruthfulQA \citep{lin-etal-2022-truthfulqa}. Adopting the same SFT training setup as in \Cref{tab: Problem Quality}, the results are summarized in \Cref{tab: Cross-Task Generalization}. 
\begin{table}[h!]
\centering
\resizebox{0.49\textwidth}{!}
{
\begin{tabular}{lccccc}
\toprule
\multirow{2}{*}{\textbf{\centering Method}} & \multirow{2}{*}{\textbf{BBH}} & \multirow{2}{*}{\parbox{1.2cm}{\centering\textbf{Human}\\ \vspace{-1mm}\centering\textbf{Eval}}}
& \multirow{2}{*}{\textbf{MMLU}} & \multicolumn{2}{c}{\textbf{TruthfulQA}} \\
 \cmidrule(lr){5-6} & & & & \small\textbf{MC1} & \small\textbf{MC2} 
\\
\midrule
Qwen2.5-7B  & 69.7 & 60.4 & 74.3 & 38.9 & 56.3\\
~~+~NuminaMath  & 67.3 & 58.5 & 73.8 & \underline{38.8} & 55.9\\
~~+~Magpie  & \underline{68.2} & 60.4 & 73.6 & 37.7 & 54.9\\
~~+~LIMO  & 67.8 & ~~\,62.2$\,^\uparrow$ & \underline{73.8} & 37.9 & \textbf{56.2}\\
~~+~S1K  & 67.4 & ~~\,\underline{63.4}$\,^\uparrow$ & 73.2 & 37.2 & 55.1\\ 
\midrule
~~+~\method & \textbf{68.4} & ~~\,\textbf{64.0}$\,^\uparrow$ & \textbf{74.1} & \textbf{38.9} & \underline{56.0}\\
\bottomrule
\end{tabular}
}
\caption{\textbf{Cross-Task Generalization Benchmarks.} \textbf{Bold} and \underline{underlined} values denote the best and second-best results among the fine-tuned models, respectively. The superscript $^\uparrow$ highlights performance improvements over the base model. TruthfulQA is evaluated using the standard MC1 (single-choice) and MC2 (multiple-choice) metrics. Our method effectively mitigates catastrophic forgetting while demonstrating positive transfer to code generation (HumanEval). 
}\label{tab: Cross-Task Generalization}
\end{table}

Overall, performance variations across datasets are marginal, highlighting two primary trends. First, a slight decline on non-coding benchmarks suggests minor forgetting, yet this effect is constrained within a narrow range, likely due to the limited scale of our training samples. Notably, our method exhibits the most robust capability retention among all fine-tuned models. Second, on the HumanEval benchmark, which shares high logical synergy with mathematical reasoning, reasoning-enhanced models including LIMO, s1K, and our approach consistently show performance gains over the base model. Among these, our method achieves the most significant improvement, demonstrating effective positive transfer to programming tasks. 



\section{Analysis}\label{sec: Dataset Analysis}
This section analyzes the characteristics of the data generated by our approach. To this end, we compare our method against the aforementioned Magpie \cite{xu2025magpie} and OpenR1-Math \cite{huggingface_openr1_2025}. Specifically, OpenR1-Math is derived from NuminaMath but serves as a rigorously filtered, high-quality subset, thus offering higher analytical value than random sampling from the raw source. 

\subsection{Data Quality and Difficulty}\label{sec: Quality and Difficulty}
\begin{figure}[t]
    \centering
    \includegraphics[width=1\linewidth]{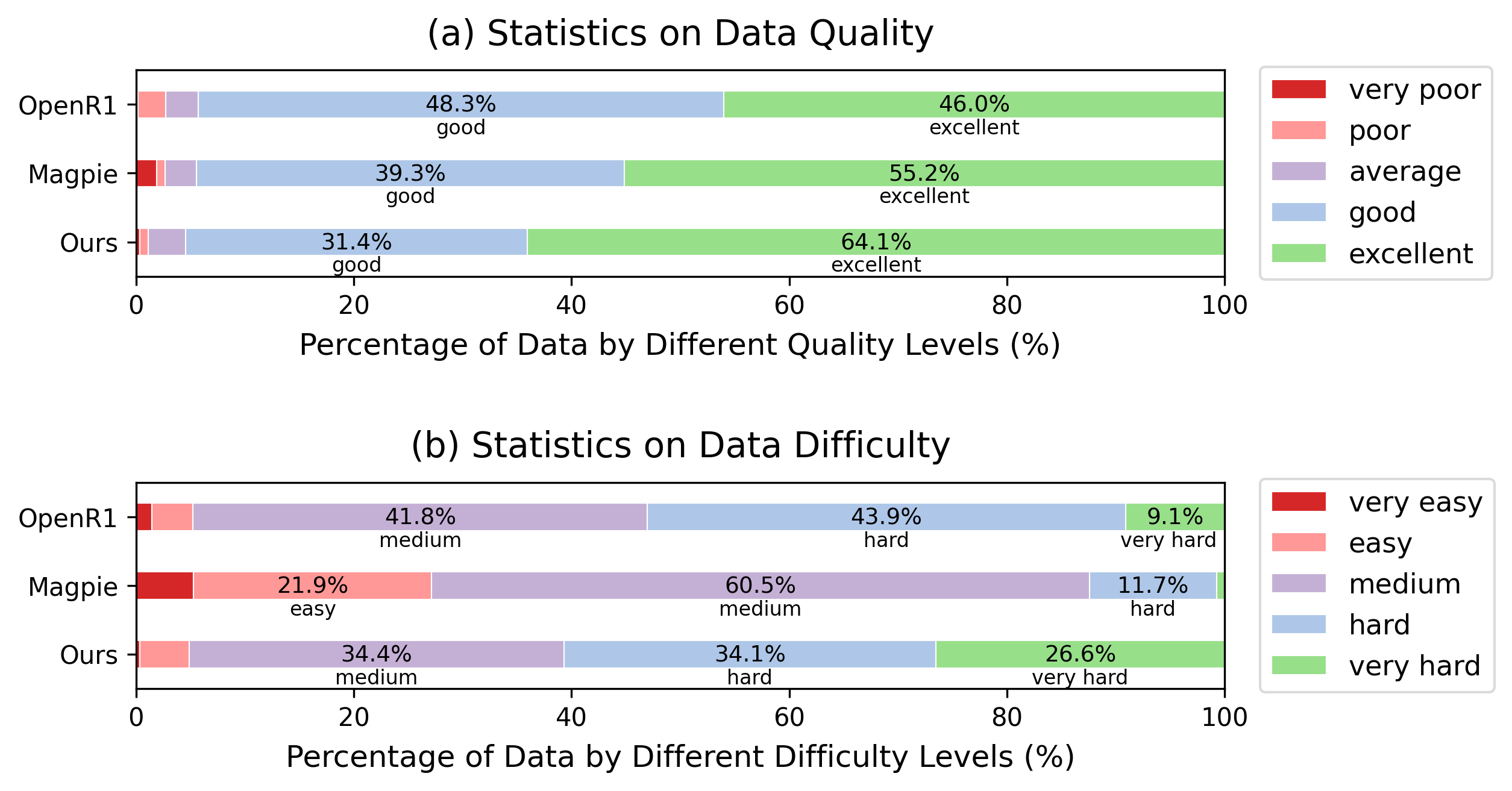}
    \caption{\textbf{Quality and Difficulty Distributions.} Quality and difficulty increase from left to right. Our method shows a significant advantage in generating high-quality, high-difficulty mathematical problems.}
    \label{fig: quality and difficulty}
\end{figure}

Following the protocol in \citet{chen2024alpagasus}, we employ \texttt{Qwen2.5-32B-Instruct} to evaluate both the quality and difficulty of the datasets on a five-level scale (The relevant prompts are detailed in Appendix \ref{sec: Prompts Analysis}). The resulting distributions are visualized in \Cref{fig: quality and difficulty}(a) and (b), respectively. For the manual evaluation of overall quality through sampling, please refer to \Cref{sec: human eval}.



\paragraph{Data Quality} As shown in \Cref{fig: quality and difficulty}(a), both synthetic datasets yield a higher proportion of "excellent" samples compared to OpenR1-Math, with our method further outperforming the Magpie baseline. While the synthetic datasets exhibit a slightly higher share of "very poor" instances than OpenR1-Math, the absolute proportion remains negligible. This is an expected consequence of the synthetic data being evaluated in its raw state, whereas OpenR1-Math has undergone post-filtering. 

\paragraph{Data Difficulty}
\Cref{fig: quality and difficulty}(b) demonstrates that our method holds a significant advantage in synthesizing high-difficulty problems, whereas Magpie is limited by its tendency to generate low-complexity, high-frequency data. In fact, our approach allows for flexible control over the adversarial iteration process via prompt engineering (e.g., style tokens). This capability enables the generation of problems at targeted difficulty levels, facilitating the construction of model-adaptive synthetic datasets.

\subsection{Dataset Diversity}\label{sec: Diversity}
Dataset diversity is widely acknowledged as a pivotal determinant of dataset quality. Despite the lack of a universal standard for its quantification, we employ two primary metrics in this study: intra-dataset similarity and dataset coverage. Intuitively, lower intra-similarity coupled with higher coverage indicates superior diversity. For experimental details of this section, please refer to Appendix \ref{section: diversity}. 

\begin{figure}[h!]
    \centering
    \includegraphics[width=1\linewidth]{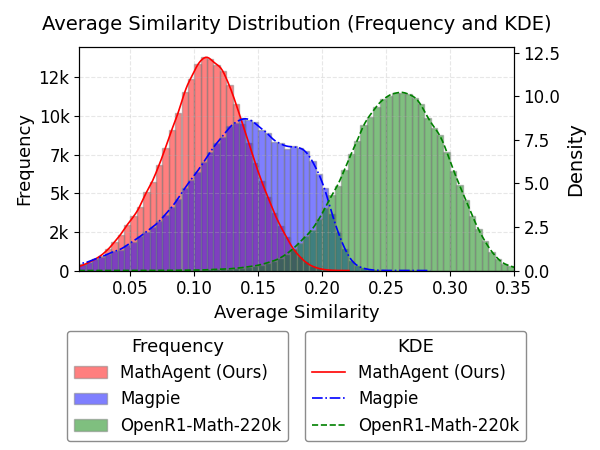}
    \caption{\textbf{Intra-dataset Similarity Distribution.} The bars represent frequency counts, while the curves denote Kernel Density Estimation (KDE). The results demonstrate that our method exhibits lower intra-similarity, indicating superior dataset diversity.}
    \label{fig: Internal_Average_Similarity}
\end{figure}

\paragraph{Dataset Intra-Similarity} 
To quantify dataset diversity, we measure intra-dataset similarity by calculating the average similarity between data instances within each dataset. We utilize the complete 220K-version OpenR1-Math dataset and randomly sample an equivalent number of instances (220K) from both our dataset and Magpie. The distribution of average similarity scores is shown in \Cref{fig: Internal_Average_Similarity}. The results demonstrate that our method achieves the lowest intra-dataset similarity.

\begin{figure}[htpb]
    \centering
    \includegraphics[width=1\linewidth]{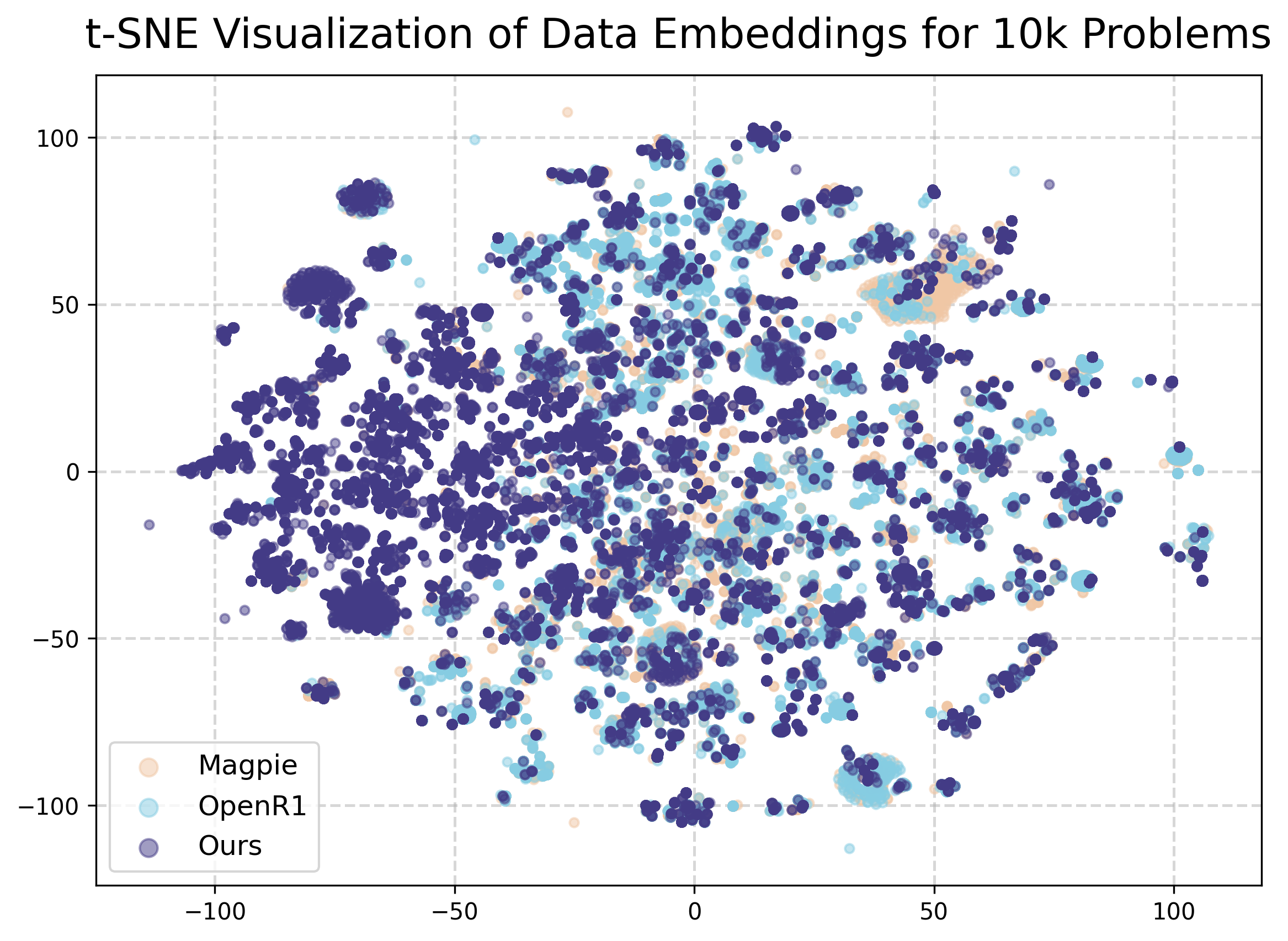}
    \caption{\textbf{t-SNE Visualization of Knowledge Points.} The extensive coverage of the dark blue points (representing our method) demonstrates the significant diversity of the generated mathematical problems.}
    \label{fig: t-SNE}
\end{figure}

\paragraph{Dataset Coverage} The coverage of mathematical problems is primarily characterized by the diversity of the underlying knowledge points. To evaluate this, we adopt an approach integrating methods from InsTag \cite{lu2024instag} and \citet{zhao2024wildchat}. 
As illustrated in \Cref{fig: t-SNE}, we employ t-SNE \citep{maaten2008visualizing} to project the semantic embeddings of knowledge point tags derived from 10K randomly sampled instances into a two-dimensional space. The results show that the distribution of the data generated by our method substantially encompasses the coverage areas of both Magpie and OpenR1-Math, which further validates the superior diversity of our approach in generating mathematical problems.

\subsection{Data Scaling}
In this section, we investigate the scaling laws governing our approach by analyzing the correlation between model performance and dataset size. We employ \texttt{Qwen2.5-Math-7B} as the base model and adjust the training schedule to 3 epochs to facilitate efficient experimentation. 

\begin{figure}[t]
    \centering
    \includegraphics[width=1\linewidth]{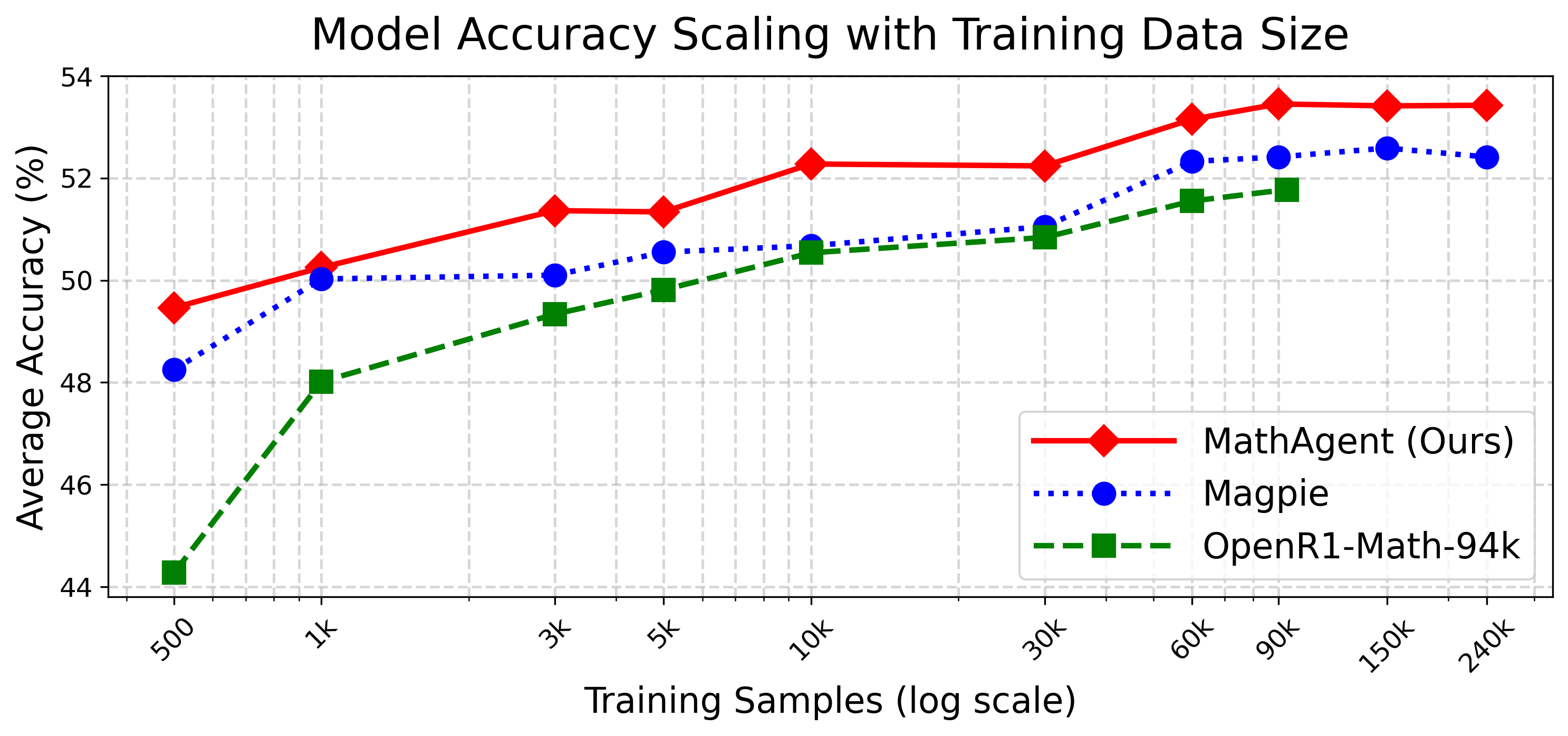}
\caption{\textbf{Performance Scaling Analysis.} The x-axis is plotted on a logarithmic scale for clarity. While performance generally improves with increased data scale, our method maintains a consistent and significant performance advantage over the baselines.}
    \label{fig: Scaling_fig}
\end{figure}

As illustrated in \Cref{fig: Scaling_fig}, performance across all three datasets exhibits a trend of rapid initial growth followed by stabilization (or slight saturation) as data scale increases, with peak performance achieved at approximately 100K samples. Crucially, our method consistently yields superior performance gains compared to other methods, demonstrating its robustness and data efficiency. 

\subsection{Human Evaluation}\label{sec: human eval}
Given the potential systematic bias of relying solely on LLM-as-a-judge, we also conduct a human evaluation. Specifically, we randomly sample 100 instances from each of Magpie, OpenR1, and \method, and perform a blinded evaluation with source anonymization on a 5-point scale, where higher scores indicate better quality. The results show that \method achieves the highest average score (4.2/5.0), outperforming Magpie (3.5/5.0) and OpenR1 (3.9/5.0).

Expert feedback further suggests that, although Magpie-generated samples are often grammatically fluent, their logical structures tend to be simplistic or repetitive, directly resulting in the lowest scores for Magpie in this evaluation. In contrast, \method demonstrates stronger structural complexity and logical depth, consistent with the high-difficulty distribution observed in the automated analysis. 
We also observe cases where the LLM judge fails to identify subtle logical flaws recognized by human experts, highlighting the limitations of automated evaluation alone. Detailed evaluation settings and discussions of limitations are provided in Appendix \ref{sec: Human Evaluation Details}.



\section{Conclusion}
By decoupling blueprint design from linguistic realization, this paper proposes the \method framework, which maximizes structural diversity through adversarial optimization on constraint graphs to guide the generation of high-quality, long-tail distributed data. Experimental results demonstrate that models fine-tuned on samples synthesized via this approach outperform existing human-filtered datasets. This provides a scalable pathway for constructing high-quality mathematical reasoning data, breaking through the limitations of manual annotation and avoiding the mode collapse issues inherent in previous synthesis methods.

\section*{Limitations}
Despite the performance gains and robust scaling characteristics demonstrated by \method, our study has several limitations:
\begin{itemize}[leftmargin=*,nosep]
\item\textbf{Domain Specificity.} Our framework currently focuses on mathematical reasoning where logic is highly structured and objective. Its applicability to more open-ended or less formal domains, such as legal reasoning or creative writing, requires further exploration into how to define effective constraint graphs for non-mathematical tasks.

\item\textbf{Computational Overhead.} The adversarial evolution process involves multiple iterations between the legislator and executor models. This iterative cycle inevitably leads to higher computational costs and longer synthesis times compared to single-pass methods that do not require multi-round refinement.

\item\textbf{Ground-Truth Verification.} As the adversarial process pushes problem complexity to extreme levels, ensuring the absolute correctness of the generated ground-truth solutions becomes increasingly difficult. Future work could benefit from integrating external formal verifiers or symbolic solvers to guarantee the accuracy of synthesized reasoning chains.
\end{itemize}
These limitations also point to potential research directions for the future.
\section*{Ethics Statement}
This work complies with the ACL Ethics Policy. We utilize publicly available datasets and open-source models, ensuring that all resources are properly cited and the experiments are reproducible. We acknowledge the inherent risks associated with LLMs, including the potential for hallucinations and the generation of non-factual content. While our method aims to enhance reasoning reliability in the mathematical domain, users should exercise caution and verify model outputs when deploying such systems in critical or real-world applications.

\section*{Acknowledgments}
The authors would like to express their sincere gratitude to the PhD students in the Department of Mathematical Sciences and the Department of Statistics and Data Science at Tsinghua University for their dedicated efforts in the manual verification process. We also thank the Huawei Large Model Data Technology Lab for its invaluable guidance and suggestions. Furthermore, we are grateful to the anonymous reviewers and the meta-reviewer for their insightful and constructive comments, which have greatly improved the quality of this work.





\bibliography{custom}
\newpage
\appendix

\section{Prompts for model training and testing}\label{sec: prompt}
Standard prompt designs in current mathematical reasoning tasks typically combine \textit{zero-shot Chain-of-Thought (CoT) guidance} with \textit{structured output constraints}. This approach strikes a balance between maintaining model accuracy and facilitating automated answer extraction and evaluation. Thus, we adopt this strategy in our study (refer to the Complex Prompt in \Cref{fig: prompt}). However, existing research indicates that such complex prompts may impose a burden on models with limited instruction-following capabilities \citep{zeng2025simplerlzoo}, potentially leading to performance degradation. Consequently, a simplified prompt design is adopted for such models (refer to the Simple Prompt in \Cref{fig: prompt}).
\begin{figure}[htpb]
\begin{tcolorbox}[
    width=0.5\textwidth,
    colback=white,
    colframe=pink,
    title=Simple Prompt,
    fonttitle=\bfseries,
    coltitle=black,
    boxrule=1pt,
    center 
]
\begin{verbatim}
Question:
{input}
Answer:
Let's think step by step.
\end{verbatim}
\end{tcolorbox}
\vspace{-4mm}
\begin{tcolorbox}[
    width=0.5\textwidth,
    colback=white,
    colframe=pink,
    title=Complex Prompt,
    fonttitle=\bfseries,
    coltitle=black,
    boxrule=1pt,
    center 
]
\begin{verbatim}
<|im_start|>system
You are a helpful assistant.<|im_end|>
<|im_start|>user
{input}
Please reason step by step, and put 
your final answer within 
\\boxed{}.<|im_end|>
<|im_start|>assistant
{output}
\end{verbatim}
\end{tcolorbox}
\caption{Comparison between simple prompts and more complex prompts (using the Qwen series prompt templates as an example).}
\label{fig: prompt}
\end{figure}

We empirically verify this phenomenon in \Cref{tab: prompt_ex}, using \texttt{Qwen3-8B-Base} and \texttt{Llama-3.1-8B} as representative models. Experimental results reveal that the non-instruction-tuned \texttt{Llama-3.1-8B} suffers significant performance degradation under complex prompts (adapted to the corresponding Llama template), whereas \texttt{Qwen3-8B-Base} demonstrates improved performance in similar settings. Based on further observations, we adopt the simple prompt for the Llama, Mistral, and Gemma series, while retaining the standard complex prompt format for the Qwen series. 
\begin{table}[htpb]
\centering
\resizebox{0.48\textwidth}{!}
{
\begin{tabular}{llc}
\toprule
\textbf{Model} & \textbf{Prompt Template} & \textbf{Avg.} \\
\midrule
\multirow{2}{*}{\parbox{3cm}{Qwen3-8B-Base\\\small{\citep{yang2025qwen3technicalreport}}}} & simple & 32.5 \\
& Qwen-complex & 49.7 \\ \midrule
\multirow{2}{*}{\parbox{3cm}{Llama-3.1-8B\\\small{\citep{grattafiori2024llama3herdmodels}}}} & simple & 10.3\\
& Llama-complex & 1.4\\
\midrule
\multirow{2}{*}{\parbox{3.5cm}{Llama-3.1-8B-Instruct\\\small{\citep{grattafiori2024llama3herdmodels}}}}  & simple & 23.5\\
& Llama-complex & 31.0 \\
\bottomrule
\end{tabular}}
\caption{Performance comparison using Simple vs. Complex prompts. Complex prompts degrade the performance of Llama-Base but benefit Qwen-Base and Llama-Instruct.}
\label{tab: prompt_ex}
\end{table}

Notably, although our experiments primarily applied complex prompt strategies to Qwen models, the applicability of this design is not limited to this series. Instead, as previously discussed, it is closely tied to the model's instruction-following capability. As shown in \Cref{tab: prompt_ex}, the instruction-tuned \texttt{Llama-3.1-8B-Instruct} also demonstrates significant performance improvement when using complex prompts (adapted to the corresponding Llama template). However, considering that instruction-tuned models are generally less suitable for few-shot supervised fine-tuning, our main experiments do not focus on them. 

\section{Training Details for SFT}\label{sec: Training Details}
We perform full-parameter fine-tuning of the model using LLaMA-Factory\footnote{\url{https://github.com/hiyouga/LLaMA-Factory}} \citep{zheng-etal-2024-llamafactory}. Adhering to the strategies discussed in \Cref{sec: prompt}, we apply the corresponding prompt templates (Simple or Complex) for each model during the data formatting stage. To optimize memory usage and training efficiency, we employ the DeepSpeed ZeRO Stage 3 \citep{rajbhandari2020zero} strategy. Training utilizes the AdamW \citep{loshchilov2018decoupled} optimizer ($\beta_1=0.9, \beta_2=0.95$, $\text{weight decay}=1 \times 10^{-4}$) combined with a cosine learning rate decay schedule \citep{loshchilov2017sgdr} and a warmup ratio of $0.05$. The maximum sequence length is limited to $4096$ tokens, and the maximum gradient norm is clipped at $1.0$. The training process uses bfloat16 precision throughout to ensure numerical stability and computational efficiency, and operates in a distributed environment across $8$ devices, with the random seed fixed at $42$ to ensure reproducibility. Certain hyperparameters, such as the initial learning rate, batch size, and number of training epochs, vary across models; these specific configurations are detailed in \Cref{tab:qwen_hyperparams}.

\begin{table}[h!]
\centering
\setlength{\tabcolsep}{1mm}
\resizebox{0.45\textwidth}{!}
{\begin{tabular}{lccc}
\toprule
\small{\textbf{Model}} & \parbox{2cm}{\centering\small{\textbf{Peak Learning}}\\\centering\small{\textbf{Rate}}} & \parbox{1.2cm}{\centering\small{\textbf{Batch}}\\ \centering\small{\textbf{Size}}} & \parbox{1.2cm}{\centering\small{\textbf{Training}}\\ \centering\small{\textbf{Epochs}}} \\
\midrule
\small{Qwen-7B/8B Series} & \multirow{2}{*}{1e-5} & \multirow{2}{*}{$2\times 8$} & \multirow{2}{*}{5} \\
\small{\& Llama-3.1-8B} \\ \midrule
\small{Qwen3-14B-Base} & \multirow{2}{*}{5e-6} & \multirow{2}{*}{$1\times 8$} & \multirow{2}{*}{5} \\ 
\small{\& Gemma-2-9B} \\ \midrule
\small{Qwen3-4B-Base} & \multirow{2}{*}{2e-5} & \multirow{2}{*}{$4\times 8$} & \multirow{2}{*}{5} \\  
\small{\& Llama-3.2-4B}\\ 
\midrule\multirow{2}{*}{\small{Mistral-7B-v0.3}} & \multirow{2}{*}{1e-5} & \multirow{2}{*}{$2\times 8$} & \multirow{2}{*}{3} \\ \\
\bottomrule
\end{tabular}}
\caption{\textbf{Training Hyperparameters} for Different Models. Note that "$\times 8$" in the batch size column denotes the total batch size calculated as \textit{per-device batch size} $\times$ \textit{number of devices} (8).}
\label{tab:qwen_hyperparams}
\end{table}

\section{Training-Test Set Similarity}\label{sec: Training-Test Set Similarity}
\begin{figure*}[t]
    \centering
    \includegraphics[width=1\linewidth]{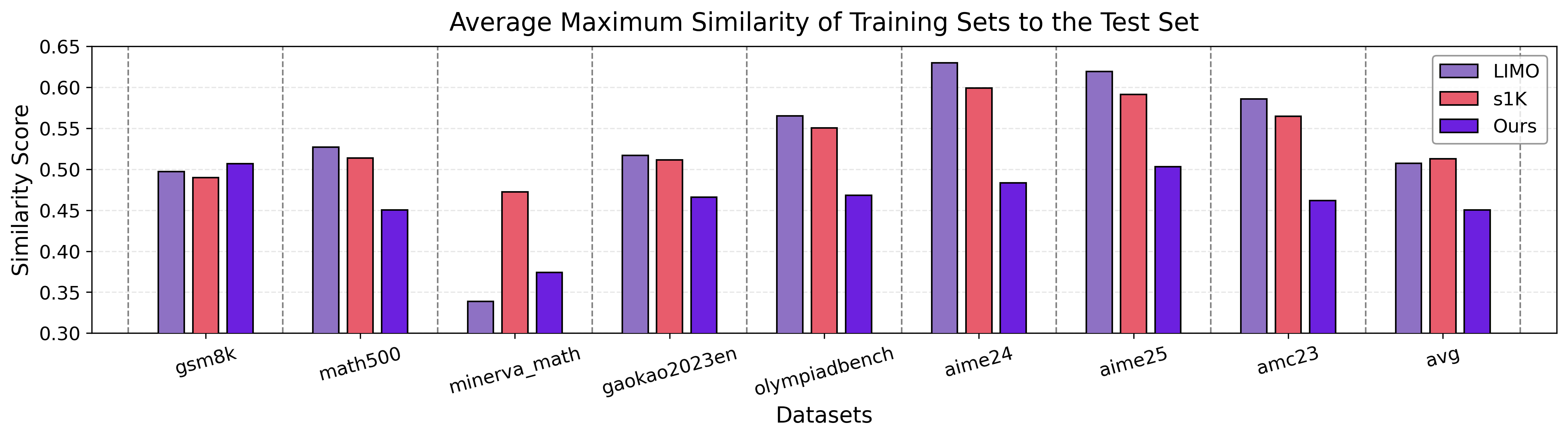}
    \caption{\textbf{Average Maximum Similarity} Between Training and Test Datasets: A lower similarity score indicates a reduced likelihood of data leakage. This figure demonstrates that our synthetic data does not carry a higher risk of data leakage compared to LIMO and s1K.}
    \label{fig: AMS_train_test}
\end{figure*}
To prevent potentially biased test results caused by data leakage during synthetic data generation, we calculated the similarity between each of the three training sets and the test set, respectively. We adopt the \textit{average maximum similarity (AMS)} in the embedding space as our metric. Specifically, we first map all mathematical problems into the embedding space using the \texttt{all-mpnet-base-v2}\footnote{\url{https://huggingface.co/sentence-transformers/all-mpnet-base-v2}} model \citep{reimers2019sentence}, which enables the computation of pairwise similarity between any two problems. For each instance in the test set, we retrieve the sample in the training set with the highest similarity to it and record this maximum similarity value. The final metric is then calculated as the average of these maximum similarity values across all test instances. 

The final results are presented in \Cref{fig: AMS_train_test}. It is evident that the similarity between our synthetic dataset and the test sets is not significantly higher than that of the other training sets. Conversely, for most test sets, the similarity metric obtained by our method is noticeably lower. This indicates that no additional information leakage occurs during our synthetic data generation process, and thus it does not introduce bias into the experimental results. 

\section{Details for Baseline Data Comparisons}
\label{sec: baseline_details}

In this section, we describe the selection criteria and implementation details of the baseline data sources used in our performance evaluation and data characteristic analysis.

\paragraph{Overall Criteria} Open-source datasets and synthesis methods are highly diverse \citep{rao2025data}. 
Because our method targets seed-free synthesis of mathematical problems, we mainly compare against methods in the same paradigm, such as Magpie \citep{xu2025magpie}, rather than seed-based approaches (e.g., Self-Instruct; \citealt{wang-etal-2023-self-instruct}). Methods such as OpenThought \citep{guha2026openthoughts}, which focus on constructing high-quality \emph{reasoning traces} from existing public datasets, are complementary to our work and therefore excluded from direct comparison. In addition, program-assisted generation frameworks such as ARROWS \citep{chen2025arrowsmathreasoningdata} are effective when problems can be formulated as executable programs, but are limited to such settings and are therefore excluded from direct comparison. Since many open-source math problems are derived from NuminaMath \citep{li2024numinamath}, we include samples from it as a raw-data baseline, together with several representative curated subsets built upon it \citep{ye2025limo,muennighoff2025s1simpletesttimescaling,huggingface_openr1_2025}.


\paragraph{LIMO\,\&\,s1K} 
We employ the open-source LIMO \citep{ye2025limo} and s1K \citep{muennighoff2025s1simpletesttimescaling} datasets as high-quality baselines. These datasets are constructed through expert-designed pipelines and rigorous screening to ensure reasoning depth, representing the state-of-the-art in small-scale, curated reasoning data. For s1K, which offers two reasoning model versions based on \texttt{Gemini} \citep{google2024gemini2} and \texttt{DeepSeek-R1} \citep{deepseekai2025deepseekr1incentivizingreasoningcapability} respectively, we specifically select the latter for the experiments in \Cref{tab: math_performance} to establish a more competitive and rigorous baseline.

\paragraph{NuminaMath} 
NuminaMath \citep{li2024numinamath} serves as a large-scale open-source collection of mathematical problems, providing a rich resource of raw mathematical tasks for research. In our experiments, we randomly sample the required number of problems from the complete dataset to evaluate the model's performance across a broad distribution of raw mathematical tasks.

\paragraph{Magpie} 
To construct the Magpie \citep{xu2025magpie} baseline, we employ \texttt{Qwen2.5-32B-Instruct} as the question generator. Following the original protocol, we adopt the math-specific prompt from Figure 2 of \citet{xu2025magpie} as the system prompt. Critically, Magpie's generation quality is highly sensitive to the sampling temperature $T$. While higher temperatures can enhance dataset diversity, they often lead to a non-negligible decline in synthesis quality; thus, we adopt a balanced setting of $T=1$ to ensure a fair and effective comparison. Unlike our iterative adversarial approach, Magpie relies on a static, uniform system prompt for direct generation. For the comprehensive data characteristic analysis, we scale this process to generate a candidate pool of 300,000 samples.

\paragraph{OpenR1-Math} 
OpenR1-Math \citep{huggingface_openr1_2025} is an open-source dataset derived from NuminaMath. As a rigorously filtered and high-quality subset, it offers significantly higher analytical value than the original raw data. The dataset consists of two versions, including a further curated set of 94k problems and a complete version containing 220k instances. For experiments that prioritize data quality over quantity, such as assessments of data quality and difficulty as well as SFT experiments for data scaling performance, we utilize OpenR1-Math-94k as the comparison baseline. In contrast, for diversity analysis, we select the larger OpenR1-Math-220k version to provide a more representative and comprehensive benchmark.

\section{Details for Diversity Analysis}\label{section: diversity}



In the internal similarity and coverage analysis, we uniformly use the \texttt{all-mpnet-base-v2} model to generate semantic embedding vectors, mapping problem descriptions and knowledge tags into a 768-dimensional vector space. The internal similarity is calculated as the average cosine similarity between the embedding vector of each sample and all other samples in the dataset pool. In the coverage analysis, the knowledge point extraction stage employs \texttt{Qwen2.5-32B-Instruct} to perform annotation on 10,000 randomly sampled mathematical problems, followed by dimensionality reduction visualization using t-SNE.

Notably, the similarity metric used here differs from that in \Cref{sec: Training-Test Set Similarity}. While the latter employs maximum similarity to detect potential data contamination from near-identical samples, the average similarity used in this section is designed to evaluate global semantic density. Furthermore, due to the high baseline similarity inherent in mathematical reasoning data, maximum similarity tends to saturate (approach 1.0) at large scales, thereby reducing its discriminative power. Consequently, we adopt average similarity to provide a more robust and distinguishable measure of dataset diversity.

\section{Human Evaluation Details}\label{sec: Human Evaluation Details}

\paragraph{Evaluation Setup} To complement the automated evaluation, we conduct an expert-based human evaluation on samples drawn from Magpie, OpenR1, and \method. We invite four Ph.D. researchers with expertise in different areas of mathematics, including Algebra, Geometry \& Analysis, Statistics, and Mathematical Physics, to ensure broad domain coverage.

We randomly sample 100 instances from each dataset, resulting in 300 evaluated instances in total. The evaluation is conducted under a blinded protocol with source anonymization, so evaluators do not know which method produced each sample. Each instance is rated on a 5-point scale, where higher scores indicate better overall quality:
\begin{itemize}[leftmargin=*,nosep]
    \item \textbf{1 (Serious Flaws):} The sample contains major logical or conceptual errors.
    \item \textbf{2 (Minor Defects):} The sample has small errors that do not completely invalidate it.
    \item \textbf{3 (Mediocre):} The sample is largely correct but trivial, repetitive, or lacking in depth.
    \item \textbf{4 (Good Quality):} The sample is correct, clear, and reasonably well-structured.
    \item \textbf{5 (Insightful):} The sample demonstrates strong logical depth, originality, or pedagogical value.
\end{itemize}

\paragraph{Additional Qualitative Findings}
Beyond the quantitative scores, the evaluators provide several consistent qualitative observations. First, Magpie-generated samples are often grammatically fluent and superficially coherent, but they frequently rely on repetitive reasoning patterns or exhibit limited structural complexity. As a result, their overall quality is often judged as moderate rather than strong.

Second, samples generated by \method are more likely to exhibit multi-step reasoning, richer structural organization, and greater logical depth. These observations are consistent with the high-difficulty tendencies identified in the automated evaluation.

Third, we observe noticeable discrepancies between human judgments and LLM-based evaluation in some cases. In particular, the LLM judge occasionally assigns favorable scores to samples that contain subtle logical gaps or weak inferential steps that are readily identified by human experts. This suggests that automated evaluation alone may fail to fully capture the quality of mathematical reasoning data.

\paragraph{Evaluation Limitations}
Despite its usefulness, the human evaluation has several limitations. First, due to the high cost and cognitive demand of expert review, the evaluation is conducted on a relatively small sample of 300 instances, which is much smaller than the full scale of the synthesized dataset.

Second, although we intentionally recruit evaluators from diverse mathematical areas, evaluator coverage remains limited relative to the full breadth of mathematical reasoning tasks. Some problems may fall closer to the expertise of certain evaluators than others. As a result, especially under limited evaluation time, some particularly difficult samples may be marked as uncertain or out-of-scope.

Third, human evaluation inevitably contains some degree of subjectivity, especially for high-level criteria such as insightfulness, elegance, or pedagogical value. While the scoring criteria help standardize judgments, they cannot completely eliminate individual variation in scoring.

\section{Prompts for Synthetic Data Generation and Analysis}\label{sec: prompts}

\subsection{Prompts for Data Synthesis}\label{sec: Prompts Synthesis}
In this section, we delineate the specific prompts employed throughout the synthetic mathematical problem generation pipeline. The hierarchical structure is organized as follows: within the \textit{Legislator} module, the prompt for the {Proposer} is detailed in \Cref{fig: Prompt Proposer}, the {Critic}'s evaluation protocol is provided in \Cref{fig: Prompt Critic}, and the {Moderator}'s decision-making logic is given in \Cref{fig: Prompt Moderator}. 

Regarding the \textit{Executor} module, we focus on its primary function, which is the {Semantic Instantiation} process. \Cref{fig: Prompt Executor} illustrates the prompt design for transforming abstract constraint graphs ($\mathcal{G}^*$) and style tokens ($\mathcal{S}$) into fluent, natural language mathematical problems. To maintain focus on our structural innovations, the subsequent solution generation and model-based verification processes are not further elaborated herein, as they follow established heuristic workflows.

\subsection{Prompts for Analysis} \label{sec: Prompts Analysis}

The prompts used for evaluating the quality and difficulty of the dataset (as discussed in \Cref{sec: Quality and Difficulty}) are designed with reference to the evaluation protocol proposed by \citet{chen2024alpagasus} and are adapted from the methodology of \citet{xu2025magpie}. The detailed prompt templates are illustrated in \Cref{fig: Prompts Quality Difficulty}.

\section{Case Study}\label{sec: Case Study}
We illustrate the core iterative process of the MathAgent framework with a representative example. For demonstration, the following style tokens are selected: \textit{Difficulty: Medium}, \textit{Question Type: Calculation}, \textit{Context: Real-world Application}, and \textit{Knowledge Level: Undergraduate}. It should be noted that although this case is very similar to the one presented in \Cref{fig: main_figure}, \Cref{fig: main_figure} primarily focuses on providing an intuitive understanding of the process. For ease of layout and process presentation, the case has been adapted accordingly.

As shown in \Cref{fig: case study} (illustrating a two-round iteration; the original JSON output has been reformatted for clarity), the $i$-th node is denoted as $v_i$, and the directed edge from node $i$ to node $j$ is denoted as $e_{ij}$. Nodes follow the ``Concept: Description'' format, conforming to the structural specifications defined in \Cref{fig: Prompt Proposer}.

\paragraph{Iteration 1} 
The initial topic is set to the \textit{Saddle Surface}, which serves as a rudimentary conceptual primitive. Driven by the Proposer ($\mathcal{A}_P$), Graph 1 is generated. The Critic ($\mathcal{A}_C$) then provides a key assessment regarding specification alignment: if the parameters $a$ and $b$ in the saddle surface equation $z = {x^2}/{a^2} - {y^2}/{b^2}$ are not constrained by $a=b$, the resulting surface integral may involve elliptic integrals, exceeding the intended \textit{Medium} difficulty. Consequently, the Critic recommends adding this constraint and further suggests including explicit nodes for partial derivatives required for surface area calculation, while leveraging symmetry to simplify the computation.

\paragraph{Iteration 2}
The Moderator ($\mathcal{A}_M$) adopts these suggestions and provides iterative guidance to the Proposer. Based on this input, the Proposer updates the graph topology to produce Graph 2. Upon reviewing Graph 2, the Critic finds no logical contradictions but offers supplementary recommendations for optimization potential:
\begin{itemize}[leftmargin=*,nosep]
    \item Introduce a node dedicated to symmetry arguments, potentially referencing group theory for theoretical depth;
    \item Incorporate dimensionless scaling analysis;
    \item Strengthen the association between $v_2$ (Context) and $v_4$ (Domain Constraint).
\end{itemize}

\paragraph{Final Decision}
The Moderator deems these additional suggestions non-essential for the current task specification, noting they might divert the problem from its core objective. Therefore, the Moderator executes \textit{adaptive truncation} to terminate the iteration. Graph 2 is output as the final \textit{Constraint Graph} ($\mathcal{G}^*$), which is then passed to the Executor for semantic instantiation into the finalized mathematical problem.

\begin{figure*}[htpb]
\begin{tcolorbox}[
    width=\textwidth,
    colback=white,
    colframe=pink,
    title=Graph 0,
    fonttitle=\bfseries,
    coltitle=black,
    boxrule=1pt,
    center 
]
$v_1$: Saddle Surface
\end{tcolorbox}
\vspace{-4mm}
\begin{tcolorbox}[
    width=\textwidth,
    colback=white,
    colframe=pink,
    title=Graph 1,
    fonttitle=\bfseries,
    coltitle=black,
    boxrule=1pt,
    center 
]
$v_1$. Saddle Surface: The abstract geometric primitive characterized by opposing curvatures.\\
$v_2$. Potato Chip Context: Real-world physical object (e.g., Pringles) used to frame the problem.\\
$v_3$. Hyperbolic Paraboloid Equation: The specific algebraic model, e.g., \(z = x^2/a^2 - y^2/b^2\).\\
$v_4$. Cylindrical Domain Constraint: The boundary condition restricting the surface to a finite disk, \(x^2 + y^2 \leq R^2\).\\
$v_5$. Surface Area Integral: The double integral calculation required to find the total area of the curved surface.\\

$e_{21}$. instantiates\\
$e_{13}$. formalized\_by\\
$e_{35}$. integrand\_source\\
$e_{45}$. defines\_limits
\end{tcolorbox}
\vspace{-4mm}
\begin{tcolorbox}[
    width=\textwidth,
    colback=white,
    colframe=pink,
    title=Graph 2,
    fonttitle=\bfseries,
    coltitle=black,
    boxrule=1pt,
    center 
]
$v_1$: Saddle Surface: The abstract geometric primitive characterized by opposing curvatures.\\
$v_2$: Potato Chip Context: Real-world physical object (e.g., Pringles) used to frame the problem.\\
$v_3$: Hyperbolic Paraboloid Equation: Simplified symmetric model: \(z = c(x^2 - y^2)\) with \(c>0\) constant, or equivalently \(a = b = 1\) in the form \(z = (x^2 - y^2)/a^2\).\\
$v_4$: Cylindrical Domain Constraint: The boundary condition restricting the surface to a finite disk, \(x^2 + y^2 \leq R^2\).\\
$v_5$: Surface Area Integral: The double integral calculation for the total area of the curved surface. Now includes symmetry exploitation: integrate over first quadrant and multiply by 4.\\
$v_6$: Partial Derivative Calculation: Compute \({\partial z}/{\partial x}\) and \({\partial z}/{\partial y}\) for the surface area element \(\mathrm{d} S = \sqrt{1 + (\partial z/\partial x)^2 + (\partial z/\partial y)^2} \, \mathrm{d}x \mathrm{d}y\).\\

$e_{21}$: instantiates\\
$e_{13}$: formalized\_by\\
$e_{36}$: differentiated\_to\\
$e_{65}$: integrand\_source\\
$e_{45}$: defines\_limits
\end{tcolorbox}
\vspace{-4mm}
\begin{tcolorbox}[
    width=\textwidth,
    colback=white,
    colframe=pink,
    title=Final Synthesized Problem,
    fonttitle=\bfseries,
    coltitle=black,
    boxrule=1pt,
    center 
]
A gourmet potato chip manufacturer designs its product to follow the precise geometry of a saddle surface, known mathematically as a hyperbolic paraboloid. When centered at the origin, the surface of a single chip is modeled by the equation $z = c(x^2 - y^2)$, where $c$ is a positive constant. To ensure uniformity, each chip is trimmed so that its vertical projection onto the $xy$-plane is bounded by the circle $x^2 + y^2 \leq R^2$.\\

By first calculating the partial derivatives ${\partial z}/{\partial x}$ and ${\partial z}/{\partial y}$ to determine the surface area element $\mathrm{d}S$, set up and evaluate a double integral to find the total surface area of the chip. In your calculation, exploit the symmetry of the surface by integrating over the first quadrant of the cylindrical domain and multiplying the result by four. Express your final answer in terms of $c$ and $R$.
\end{tcolorbox}
\caption{Case Study.}
\label{fig: case study}
\end{figure*}


\begin{figure*}[htpb]
\begin{tcolorbox}[colback=black!5!white, colframe=black!75!black, 
     title=Prompt for Legislator - Proposer ($\mathcal{A}_P$)]
You are the Proposer ($\mathcal{A}_P$) in the Legislator-Executor framework. Your objective is to drive the meta-level structural evolution of a mathematical problem by optimizing a Constraint Graph $\mathcal{G} = (\mathcal{V}, \mathcal{E})$. \\\\
\#\#\# Input Data:\\
- Style Tokens ($\mathcal{S}$): \{{STYLE\_TOKENS\_INPUT}\} \\
- Current Graph ($\mathcal{G}_t$): \{{CURRENT\_GRAPH\_DATA}\} \\
- Feedback: \{{ITERATIVE\_GUIDANCE\_FROM\_MODERATOR}\} \\\\
\#\#\# Operational Directives:\\
- Evolution \& Revision: Perform topological mutations to transition $\mathcal{G}_t$ to $\mathcal{G}_{t+1}$. \\
- Graph-Style Alignment: Expand nodes $\mathcal{V}$ and logical edges $\mathcal{E}$ to achieve the graph-related stylistic goals (particularly complexity) specified in $\mathcal{S}$. \\
- Consistency Maintenance: Rectify any logical contradictions identified in previous feedback. \\\\
\#\#\# Task Workflow:\\
Step 1: Internal Analysis \& Planning \\
Analyze the gap between the current graph $\mathcal{G}_t$ and the target specifications in $\mathcal{S}$. Plan specific mutations (e.g., adding concepts, nesting operators, or refining constraints) to bridge this gap while resolving any reported flaws. \\
Step 2: Structured Output (JSON) \\
Generate the updated graph $\mathcal{G}_{t+1}$ following the strict JSON schema below. \\\\
\#\#\# Final Output Format:\\
Analysis and Planning: [Your detailed step-by-step thinking process here] \\
Final Optimized Graph (JSON):
\begin{verbatim}
{
  "graph_id": "G_{t+1}",
  "nodes": [{"id": "v_n", "concept": "string", "description": "string"}],
  "edges": [{"source": "v_i", "target": "v_j", "relation": "string"}],
  "mutation_log": "Summary of changes made in this iteration."
}
\end{verbatim}
(Constraint: Ensure all referenced nodes in `edges` exist in the `nodes` list)
\end{tcolorbox}
\caption{Prompt for the Proposer.}\label{fig: Prompt Proposer}
\end{figure*}

\begin{figure*}[htpb]

\begin{tcolorbox}[colback=black!5!white, colframe=black!75!black, 
     title=Prompt for Legislator - Critic ($\mathcal{A}_C$)]
You are the Critic ($\mathcal{A}_C$). Your goal is not merely to check for correctness, but to identify the "evolutionary headroom" of the graph $\mathcal{G}_{t+1}$ to push it from functional to exceptional.  \\\\
\#\#\# Input for Review:\\
- Style Tokens ($\mathcal{S}$): \{{STYLE\_TOKENS\_INPUT}\}\\
- Proposed Graph ($\mathcal{G}_{t+1}$): \{{PROPOSED\_GRAPH\_DATA}\}\\\\
\#\#\# Evaluation Dimensions:\\
- Internal Consistency: Scrutinize for logical contradictions or ill-defined constraints.\\
- Specification Alignment: Verify if the graph strictly complies with the complexity and category requirements in $\mathcal{S}$.\\
- Optimization Potential: Even if requirements are met, provide several actionable suggestions for potential optimization.
\\\\
\#\#\# Final Output Format:\\
- Analysis: [Your detailed step-by-step thinking process here]\\
- Critical Flaws: [List any issues that violate consistency or $\mathcal{S}$ (Output "None" if perfect)]\\
- Refinement Suggestions: [Propose at least 2-3 specific actions to further optimize the graph's complexity or elegance]\\
- Expected Utility: [Estimate the marginal gain of these optimizations (High\,/\,Medium\,/\,Low) to assist the Moderator's decision]

\end{tcolorbox}\caption{Prompt for the Critic.}\label{fig: Prompt Critic}
\end{figure*} 

\begin{figure*} 
\begin{tcolorbox}[colback=black!5!white, colframe=black!75!black, 
     title=Prompt for Legislator - Moderator ($\mathcal{A}_M$)]
You are the Moderator ($\mathcal{A}_M$). You adjudicate the state of graph $\mathcal{G}_{t+1}$ based on the Critic's report. \\\\
\#\#\# Data for Decision:\\
- Critic's Report: \{{CRITIC\_REPORT}\}\\
- Style Tokens ($\mathcal{S}$): \{{STYLE\_TOKENS\_INPUT}\}\\
- Proposed Graph ($\mathcal{G}_{t+1}$): \{{PROPOSED\_GRAPH\_DATA}\}\\\\
\#\#\# Decision Logic:\\
- Adaptive Truncation: If $\mathcal{G}_{t+1}$ satisfies $\mathcal{S}$ and the potential for further gain is marginal, $\mathcal{A}_M$ terminates the process and outputs the graph.\\
- Iterative Guidance: Otherwise, direct specific modifications to the Proposer to extend structure or rectify flaws.\\\\
\#\#\# Final Output Format:\\
- Analysis: [Your detailed step-by-step thinking process here]\\
- Decision: [Suspend/Continue Iteration]\\
- Guidance for the Proposer: [If ITERATE: Provide a concise instruction list for the Proposer. If TERMINATE: Output "None"]\\
- Final Graph: [If TERMINATE: Output the full JSON of $\mathcal{G}_{t+1}$. If ITERATE: Output "N/A"]
\end{tcolorbox}
\caption{Prompt for the Moderator}\label{fig: Prompt Moderator}
\end{figure*}

\begin{figure*}[htpb]\begin{tcolorbox}[colback=blue!2!white, colframe=blue!75!black,title=Prompt for Executor - Question Synthesizer]
Your task is to perform Semantic Instantiation: converting an abstract Constraint Graph $\mathcal{G}^*$ into a high-quality, natural language mathematical problem.\\\\
\#\#\# Input Data: \\
- Style Tokens ($\mathcal{S}$): \{{STYLE\_TOKENS\_INPUT}\}\\
- Final Constraint Graph ($\mathcal{G}^*$): \{{FINAL\_GRAPH\_DATA}\}\\\\
\#\#\# Operational Directives: \\
- Structural Fidelity: Every node $v \in \mathcal{V}$ and edge $e \in \mathcal{E}$ must be reflected in the problem. Do not omit constraints.\\
- Style Alignment: The generated mathematical problem should conform to the constraints specified in the style tokens.\\
- Semantic Fluency: The problem must be linguistically fluid, not a robotic list of conditions. Ensure logical transitions between the situational narrative and the technical specifications.\\
- Output Constraint: Generate ONLY the natural language question ($Q$). Do not provide solutions, explanations, or meta-comments.\\\\
\#\#\# Final Output Format:\\
- Analysis:  [Step-by-step plan: How to map $\mathcal{G}^*$ nodes to $\mathcal{S}$ context while maintaining fluency]\\
- Question: [The finalized natural language problem statement]
\end{tcolorbox}\caption{Prompt for the Question Synthesizer.}\label{fig: Prompt Executor}\end{figure*}

\begin{figure*}[htpb]
\begin{tcolorbox}[colback=cyan!5!white, colframe=cyan!75!black,  title=Prompt for Generating Quality of Problems,]
\# Instruction\\
You need to rate the quality of the math problem based on its clarity, accuracy, and logical coherence.
The rating scale is as follows:\\

\begin{itemize}[label=--,leftmargin=*,nosep]
    \item \textbf{Very poor:} The problem description is ambiguous, conditions are incomplete, or contains logical contradictions. It lacks essential information and context required for solving, or the given instruction is not a mathematical problem.
    
    \item \textbf{Poor:} The problem is somewhat unclear or lacks important details. It requires significant clarification to define the solving requirements.
    
    \item \textbf{Average:} The problem is moderately clear and accurate but may contain imprecise expressions. Additional information might be needed for a complete solution.
    
    \item \textbf{Good:} The problem is clearly structured, with well-defined conditions and logical coherence. It provides sufficient information to support the solving process.
    
    \item \textbf{Excellent:} The problem is precisely formulated, with complete conditions and rigorous logic. It contains all necessary elements for solving without redundant information.
\end{itemize}

~

\#\# Math Problem to Evaluate\\
\{math\_problem\}\\

\#\# Output Format \\
First, provide an assessment highlighting the strengths and/or weaknesses of the math problem. Then, output a rating by filling in the placeholders:\\

"explanation": "[Your assessment analysis]", \\
"quality": "[very poor\,/\,poor\,/\,average\,/\,good\,/\,excellent]".
\end{tcolorbox}
\begin{tcolorbox}[colback=pink!5!white, colframe=pink!75!black, 
    title=Prompt for Generating Difficulty of Problems,]
\# Instruction\\
You are an expert in mathematics education and cognitive task analysis. Your responsibility is to evaluate the complexity of mathematical problems presented by users. For each mathematical problem, you must first identify the required knowledge points, and then assess the difficulty level based on the mathematical concepts involved, problem-solving steps, and cognitive demands.\\

\#\# Math Problem to Evaluate\\
\{math\_problem\}\\

\#\# Output Format\\
Given the provided mathematical problem, in your output you must first determine the knowledge points required to solve it. Then, rate the difficulty level of the mathematical problem as 'very easy', 'easy', 'medium', 'hard', or 'very hard'.\\

Please output the difficulty level below in the following format by filling in the placeholders in [...]:\\

"explanation": "[Your detailed explanation and reasoning]",\\
"knowledge": "[list specific mathematical concepts, procedures, or knowledge domains]",\\
"difficulty": "[very easy\,/\,easy\,/\,medium\,/\,hard\,/\,very hard]".
\end{tcolorbox}
\caption{Prompts for Evaluating Mathematical Problem Quality and Difficulty.}\label{fig: Prompts Quality Difficulty}
\end{figure*}

\end{document}